\documentclass[letterpaper]{article} 
\usepackage{aaai23}  
\usepackage{times}  
\usepackage{helvet}  
\usepackage{courier}  
\usepackage[hyphens]{url}  
\usepackage{graphicx} 
\usepackage{natbib}  
\usepackage{caption} 
\usepackage{algorithm}
\usepackage{algorithmic}
\usepackage[switch]{lineno} 
\usepackage{newfloat}
\usepackage{listings}
\DeclareCaptionStyle{ruled}{labelfont=normalfont,labelsep=colon,strut=off} 
\floatstyle{ruled}
\newfloat{listing}{tb}{lst}{}
\floatname{listing}{Listing}
\usepackage{subfigure}
\usepackage{multirow}
\usepackage{amssymb}
\usepackage{amsmath}
\usepackage{booktabs}

\title{Learning Continuous Depth Representation via Geometric Spatial Aggregator}
\author {
    Xiaohang Wang\equalcontrib,\textsuperscript{\rm 1}
    Xuanhong Chen\equalcontrib,\textsuperscript{\rm 1}
    Bingbing Ni,\textsuperscript{\rm 1}\thanks{Corresponding author: Bingbing Ni.}
    Zhengyan Tong,\textsuperscript{\rm 1}
    Hang Wang,\textsuperscript{\rm 1}
}
\affiliations {
    \textsuperscript{\rm 1} Shanghai Jiao Tong University, Shanghai 200240, China \\
    \{xygz2014010003, chen19910528, nibingbing, 418004\}@sjtu.edu.cn,  francis970625@gmail.com
}

\begin{document}
\maketitle

\begin{abstract}
Depth map super-resolution (DSR) has been a fundamental task for 3D computer vision. 
While arbitrary scale DSR is a more realistic setting in this scenario, previous approaches predominantly suffer from the issue of inefficient real-numbered scale upsampling.
To explicitly address this issue, we propose a novel continuous depth representation for DSR.
The heart of this representation is our proposed Geometric Spatial Aggregator (GSA), which exploits a distance field modulated by arbitrarily upsampled target gridding, through which the geometric information is explicitly introduced into feature aggregation and target generation.
Furthermore, bricking with GSA, we present a transformer-style backbone named GeoDSR, which possesses a principled way to construct the functional mapping between local coordinates and the high-resolution output results, empowering our model with the advantage of arbitrary shape transformation ready to help diverse zooming demand.
Extensive experimental results on standard depth map benchmarks, e.g., NYU v2, have demonstrated that the proposed framework achieves significant restoration gain in arbitrary scale depth map super-resolution compared with the prior art. 
Our codes are available at \url{https://github.com/nana01219/GeoDSR}.

\end{abstract}

\section{Introduction}

Depth maps have been used to assist in solving many complex computer vision problems, such as SLAM~\cite{tateno2017cnn,cui2020sdf}, semantic segmentation~\cite{weder2020routedfusion,gupta2014learning}, and 3D reconstruction~\cite{choe2021volumefusion,chen20203d}, etc.
However, due to budget and low-power considerations, the resolution of modern depth sensors on consumer devices is often much lower than the corresponding RGB images, limiting the application feasibility of the depth modality. Therefore, an emerging research topic is to upsample a low-resolution (LR) depth map to high-resolution (HR), guided by the corresponding high-resolution RGB images.

Most recent CNN-based studies have focused on improving super-resolution results at fixed integer scales (e.g., $\times2/4/8/16$). However, it is inconsistent with real-world application requirements, where the wanted scaling factors of the given scenes are usually variable real numbers or even unknown. Thus, we argue that scale-continuous upsampling is more practical for RGB-guided depth super-resolution, e.g., gradual zooming of the depth map when shooting. 
Additionally, in some user-specified scenarios, the shape of the collected depth map may be warped, such as wide-angle panoramic scanning (e.g., LiDAR), where the detection region is an arc-shaped curved surface. In these cases, spatial-continuous upsampling is needed to transform an input depth map to the desired geometry.
All these demands a continuous representation of the depth map, through which real-numbered scale spatial feature upsampling should be enabled, and the shape of sampling regions can be arbitrarily changed.


\begin{figure}[t]
\centering
\subfigure[Scale-continuous sampling]{
    \includegraphics[width=0.97\columnwidth]{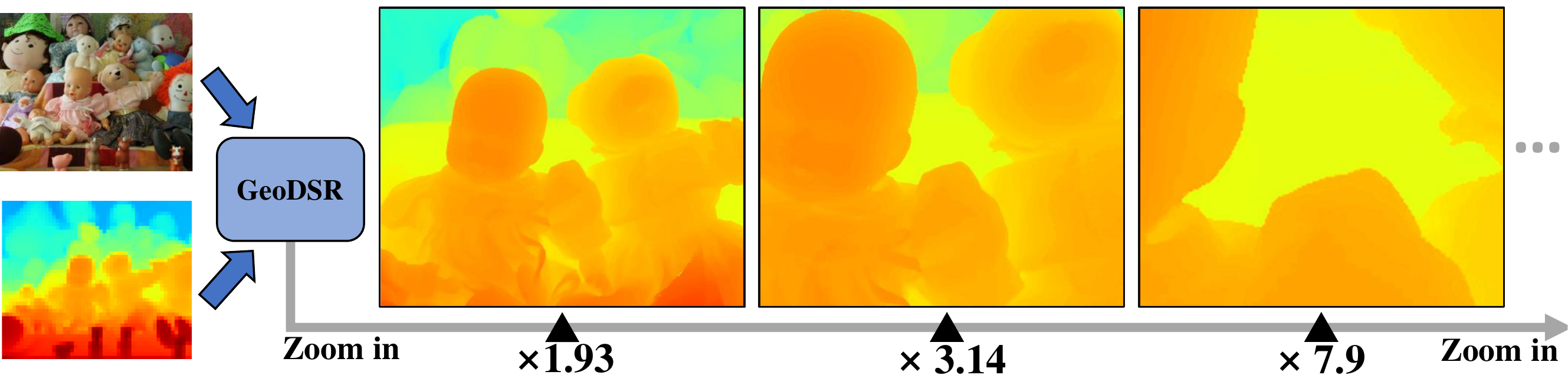}
}
\subfigure[Spatial-continuous sampling]{
    \includegraphics[width=0.97\columnwidth]{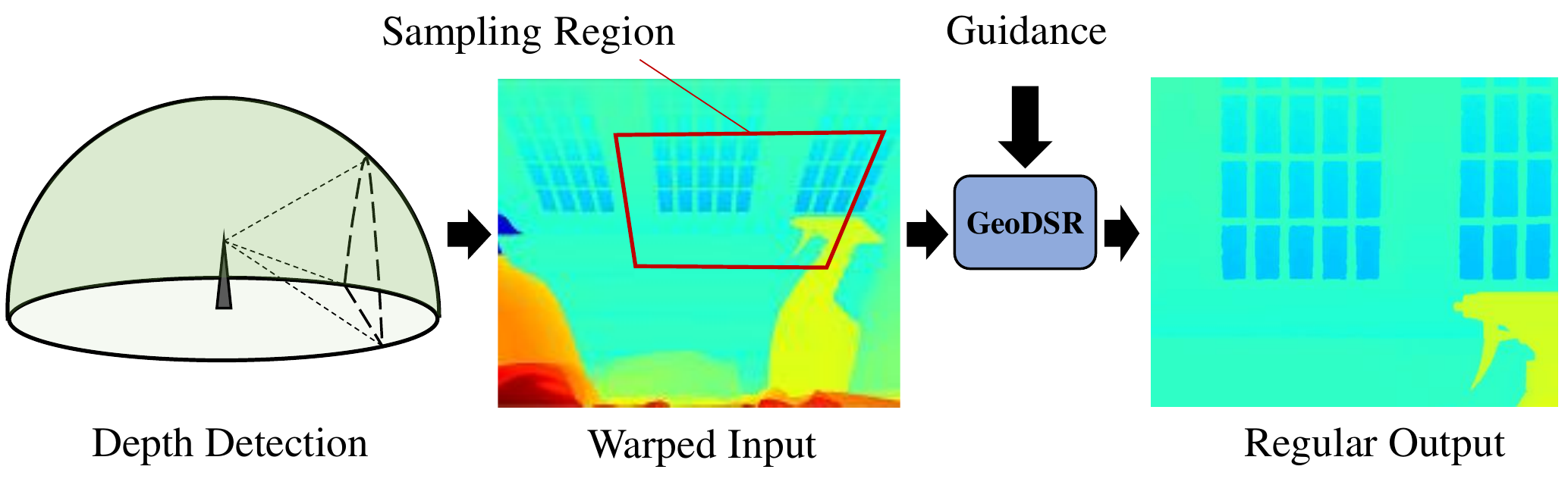}
}
\caption{Construct continuous representation for depth maps with GSA. (a) Our continuous representation can be sampled with arbitrary sampling ratios/scales while maintaining high fidelity. (b) Sample a specific area or shape (e.g., scanned maps with LiDAR) in our continuous representation by specifying the query coordinates to achieve the effect of image warping.}
\label{fig:intro}
\end{figure}


We need to pay special attention to three key points to learn such a continuous representation of depth with neural networks. 
First, \emph{scale continuity.} The model must be able to effectively integrate the geometric information of the depth map and generate a continuous feature space that can be dynamically changed to accommodate different scaling factors.
Second, \emph{spatial continuity}. The sampling method should be flexible enough to gracefully sample arbitrarily shaped areas of the depth map.
Third, \emph{efficient information aggregation}. As a multi-modal task, how to efficiently fuse depth and guidance information must be considered when constructing a continuous representation for DSR.
Prior works such as~\cite{metaSR,Scale-aware} have existed on upsampling at arbitrary resolutions in the RGB domain. However, most of these approaches are scale-based and difficult to extend to depth continuous representation.~\cite{Ye2020DepthSV} proposed a multi-scale learning method for depth super-resolution, but their model is specially conducted on several discrete scales, which is still far from true continuity.

In order to explicitly address these issues, we propose a novel transformer-style framework named \emph{GeoDSR} to construct a high-performance spatial/scale continuous representation for RGB-guided depth super-resolution.
To gain the spatial/scale continuous adaptive capability, we present the~\emph{Geometric Spatial Aggregator (GSA)} as the computational primitive, which extracts the texture pattern in a bilateral way (i.e., mean value of Gaussian process~\cite{damianou2013deep}).
In detail, the GSA consists of two modeling components: the geometric encoder and the static learnable kernel.
Analogous to the Gaussian process, the geometric encoder utilizes the scale modulated distance field to establish the correlation between sampling points, endowing GSA scale and spatial adaptability.
The static part is responsible for learning common texture pattern processing knowledge from the training data, which ensures the basic processing capability of our operator.
Furthermore, to fully utilize RGB and depth information, we design an extremely simple and efficient feature modulation operation for information fusion, which directly multiplies and regresses the fully extracted RGB and depth features.
Compared with other effortless fusion methods such as concatenation and AdaIN~\cite{huang2017arbitrary}, feature modulation can explicitly enhance the features (i.e., crucial edges) shared by RGB and depth map by using multiplication operations while suppressing noise (i.e., undesired textures).
With the seamless cooperation of the above parts, as shown in Figure~\ref{fig:intro}, GeoDSR can effectively handle both scale-continuous and spatial-continuous upsampling.
To the best of our knowledge, our GeoDSR network is the first work to learn continuous representation for depth maps.

Summarily, the contributions of this paper are as follows:

(1) We propose a novel framework (i.e., GeoDSR) and operator (i.e., GSA), which together achieve the ability to aggregate geometric information and learn the continuous representation of depth maps;

(2) Through the continuous representation, our model can effectively realize both scale-continuous and spatial-continuous upsampling in guided depth super-resolution;

(3) Extensive experiments show that our proposed continuous method outperforms state-of-the-art results on main depth SR benchmarks even compared to the models trained with fixed upsampling scale factors.

\section{Related Works}

\subsubsection{Guided Depth Super-Resolution}

Previous methods on guided depth map super-resolution can be mainly divided into the following three categories according to their problem-solving ideas.
\emph{Filter-based methods}~\cite{kopf2007joint,zhang2014rolling,ma2013constant,he2012guided} focus on constructing a joint filter whose parameters are determined by the guide image.
\emph{Optimization-based methods}~\cite{diebel2005application,park2011high,ferstl2013image,yang2012depth} regard the entire task as an optimization problem, and seek super-resolution results by minimizing the value of the corresponding optimization function.
\emph{Learning-based methods}~\cite{xu2015deep,DMSG2016,DJFR,Gu_2017_CVPR,he2021towards} often rely on a large amount of data to train a model that can extract more useful features from RGB images, so as to achieve better results.
Recently, with the increasing learning ability of neural networks, many learning-based approaches~\cite{IJCV2021,JIIF,ye2020pmbanet,he2021towards,AHMF,DCTNet} have achieved high-quality results.
However, most of these works only focus on improving super-resolution results at fixed scales, which is inconsistent with actual demand.
\section{Methodology}

\begin{figure*}[t]
\centering
\includegraphics[width=1\textwidth]{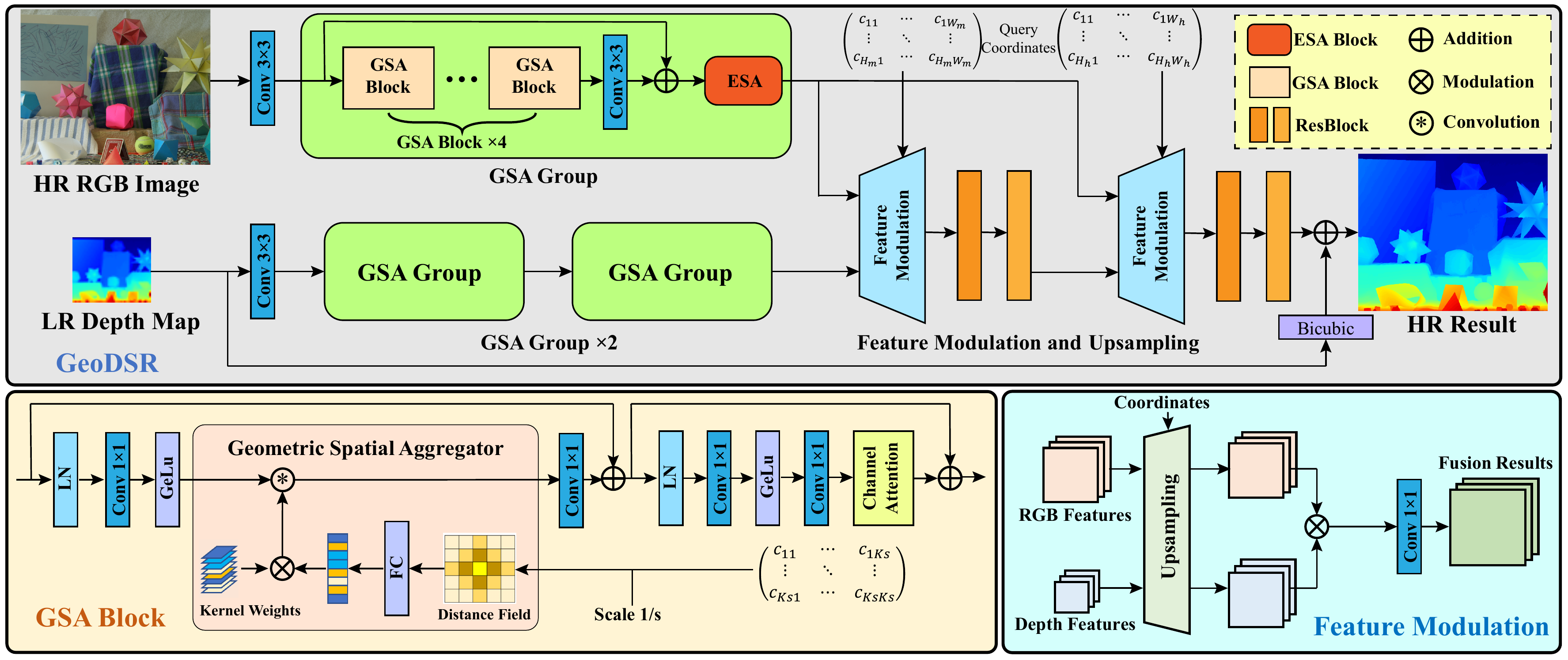}
\caption{The structure of the proposed GeoDSR network. The upper subfigure shows the overall framework of the network, and the lower two parts show the structures of the GSA Block and feature modulation module, respectively.  }
\label{fig:framework}
\end{figure*}

In this section, we first formulate the problem and demonstrate why continuous representations are essential for guided depth map super-resolution. Then we present our proposed key continuous-domain learning component, called \emph{Geometric Spatial Aggregator} (GSA), which acts as a scale perceptual joint inference module during the feature extraction process. We also build a multi-modal upsampling module, which proves to utilize the spatial geometry information extracted by GSA efficiently. At last, we introduce the overall framework of our GeoDSR network, which is shown in Figure~\ref{fig:framework}.

\subsection{Problem Formulation Definition}  

Our problem setup is RGB-guided depth super-resolution. Namely, given a LR depth map $D_l \in \mathbb{R}^{h\times w\times 1}$ and its corresponding HR RGB image $Y_h \in \mathbb{R}^{H\times W\times 3}$, the task is to upsample $D_l$ under the guidance of $Y_h$ to get the HR depth map $D_h \in \mathbb{R}^{H\times W\times 1}$. This can be formulated as:

\begin{equation}
  D_h = \Psi_\theta(D_l, Y_h; s),
\end{equation}
where, $s \in \mathbb{R}$ denotes the upsampling ratio/scale, which is calculated by $H/h$ or $W/w$; and $\Psi(\cdot)$ denotes the field function, usually a learnable model with weights $\theta$. Existing DSR frameworks~\cite{Gu_2017_CVPR,he2021towards,IJCV2021,AHMF} mainly focus on fixed and discrete upsampling scales, e.g., $\times2/4/16$. However, in practical applications, the resolutions (i.e., $H, W, h, w$) of the obtained RGB images and depth maps are not fixed due to different hardware parameters of devices. As a result, the required scale $s$ is variable and probably not an integer.

Unlike these fixed scale-based models, continuous SR methods attempt to upsample the input image/map according to real-valued coordinates. 
This process could be regarded as a coordinate-to-value mapping function: given a particular input coordinate, the signal intensity at that location will be predicted as the function output. Taking $c_{i,j}$ as the querying coordinate, the function can be formulated as:

\begin{equation}
  D_h(c_{i,j}) = \Psi_\theta(D_l, Y_h; s, c_{i,j}).
\end{equation}

\subsection{Geometric Context Learning}
Obviously, when sampling of arbitrary resolution/shape, how to effectively utilize the correlations among sampling points is crucial. Our method is inspired by the Gaussian process, which usually uses the predicted mean of Gaussian distribution to model the target to be predicted and consists of a dynamic part related to distance and a static part related to training data. When the dynamic part is applied to interpolation to represent the function of distance and correlation, it is usually modeled with a prior model, such as RBF, Laplacian, dynamic convolution~\cite{DBLP:journals/corr/abs-2106-02253,DBLP:conf/aaai/HuCN0L22} etc. 
Correspondingly, our GSA operator also includes a dynamic part of modeling distance correlation composed of distance field derived from sampling coordinates and FC module, and a knowledge part of dataset composed of static convolution operator.
Compared with the traditional Gaussian process, our operator has more powerful learning components and can perform hierarchical stacking and nonlinear filtering, which enables our operator to handle more complex texture patterns. Compared with the operators of heuristic design, our GSA has more explicit interpretability.

\subsection{Geometric Spatial Aggregator}
As illustrated in the ''GSA Block'' part of Figure~\ref{fig:framework}, our GSA operator consists of two inputs: features from previous networks and the distance field produced by sampling coordinates. 
The former is the prior knowledge learned from the data, while the latter is the spatial geometric information that changes dynamically according to the sampling conditions.
As mentioned above, in the Gaussian process, a prior distribution such as RBF or Laplacian kernel is usually applied to model the correlation in the distance. 
However, these kernels generally do not have enough learnable parameters, and their fitting ability under complex situations is limited. Thus, we use a FC module to replace the prior model, enhancing the aggregator's capability of dynamically extracting features according to the geometric information between the sampling points.
If $c$ and $c'$ represent the index positions of two points in the coordinate matrix, then the distance correlation calculation between them can be expressed as:

\begin{equation}
        \mathcal{W}^g(c, c'; s) = \Phi \left(\frac{1}{s\cdot \|c-c' \|_2^2}\right),
\end{equation}
where $s$ indicates the scale factor and $\Phi$ denotes the learnable FC module.

To mix geometric information and prior knowledge, we map the distance correlation into a vector which will be modulated to the convolution kernel in the backbone network. Thus, the whole process can be formulated as:

\begin{equation}
\small
\begin{split}
\mathcal{A}^d_{i,j}(X;c_{i,j}) = \sum_{(u, v)\in \Delta_{i, j}} \overbrace{\mathcal{W}^s_{u,v}}^{Static}\cdot \overbrace{\mathcal{W}^{g}(c_{i,j},c_{u,v})}^{Dynamic}\ast X_{u,v},
\end{split}
\end{equation}
where $\mathcal{A}^d_{i,j}$ denotes the output value of GSA at $c_{i,j}$, $X$ denotes the input feature, $W^s$ denotes the static learnable kernel, $\ast$ denotes convolution, and
{\small $\Delta_{i,j}$} denotes the set of positions in the {\small $K_s \times K_s$} kernel window, written under Cartesian product as

\begin{equation}
\small
\begin{split}
\Delta_{i,j} = [0, 1, \cdots, 2\lfloor K_s/2 \rfloor ] \times [0, 1, \cdots, 2\lfloor K_s/2 \rfloor ].
\end{split}
\end{equation}
In our framework $K_s = 3$.


\subsection{Feature modulation module}
Traditional learning-based DSR methods~\cite{JIIF,IJCV2021,DJFR,DMSG2016} often directly concatenate the features of guided images and depth maps and feed them into the neural network as different channels. Since these features are still learned separately in essence, therefore, the model requires additional resources to achieve both modality fusion. For the sake of lighter and more efficient fusion, we design a simple yet effective pre-fusion module. As shown in Figure~\ref{fig:framework}, this module only contains a grid-based upsampler, a modulation operator, and a $1\times1$ convolution layer. The input features will be firstly upsampled to the corresponding shape according to the coordinate matrix, and the subsequent fusion process at a query coordinate $c_{i,j}$ can be formulated as:

\begin{equation}
    X_{i,j} = \omega[X_{Y_h}(c_{i,j}) \otimes X_{D_l}(c_{i,j})]+b,
\end{equation}
where $X_{Y_h}$ and $X_{D_l}$ denote the upsampled feature maps of the HR RGB image and LR depth image, respectively, $\omega$ and $b$ are weights of the convolution, and $\otimes$ denotes a modulation operation.

Modulation is similar to an ``AND" operation; namely, the output signal where both input channels are strong gets amplified, while the signal where either channel is weak gets attenuated. This characteristic is helpful for pruning extraneous textural regions in object surfaces and enhancing natural object boundaries in the depth domain. Compared with directly concatenating, modulation allows pre-fusion of features, improving learning efficiency and interpretability.

\subsection{Network Scheme and Learning Objective}  

\subsubsection{Overall Framework}
Figure~\ref{fig:framework} shows our overall network architecture. 
Following the current trend, we use the transformer structure (i.e., replace the self-attention module in the traditional transformer block with our GSA operator) to build a GSA Block.
By connecting several GSA Blocks and one Enhanced Spatial Attention (ESA)~\cite{ESA} module in series, we get a transformer-style feature extraction structure, GSA Group, in which we add a residual structure to ensure smooth gradient propagation. 
Considering the balance of the performance and efficiency, when extracting the features of the guide image, we use one GSA Group, while in the depth branch, we use two. 


In the subsequent network structure, we combine the feature modulation module and ResBlock as the feature decoder.
We adopt a progressive upsampling strategy based on two decoders, where the first one enlarges the depth map to an intermediate resolution, which is then up-scaled to the target resolution by the second one. Both fusion modules receive guidance information for RGB pictures. There are two main reasons for this design: first, compared with using only one decoder, the actual magnification of each decoder under this strategy is smaller, which reduces the difficulty of upsampling, especially when the magnification is large. Second, the model has two fusion modules in this situation, which can prevent the loss of guided image information in the deep layers of the network. The intermediate resolution is default set to the arithmetic mean of the initial and target resolutions. 



\subsubsection{Loss Function} We train our model by minimizing the L1-loss between the predicted result and the ground truth as follows:

\begin{equation}
    \mathcal{L}_1(\widehat{D}_{h}, D_{h}) = \frac{1}{N} \sum_{i=1}^{N} \left \|\widehat{D}_{h(i)} -D_{h(i)}  \right \| _1,
\end{equation}
where $\widehat{D}_{h(i)}$ is the output of our model, $D_{h(i)}$ denotes the ground truth, and $N$ denotes the set of sampling points.

\setlength{\tabcolsep}{5pt}
\begin{table*}[t]
\begin{center}
\begin{tabular}{c|c|ccc|ccc|ccc}
\toprule[ 2pt]
 \multirow{2}{*}{Method}  & \multirow{2}{*}{continuous} 
 & \multicolumn{3}{c|}{NYU v2}
 & \multicolumn{3}{c|}{Middleburry}                                          & \multicolumn{3}{c}{Lu}                                                                                                    \\ \cline{3-11} 
                          &  & $\times4$ & $\times8$ &  $\times16$ & $\times4$ & $\times8$ & $\times16$ & $\times4$ & $\times8$ & $\times16$ \\ \hline
\multicolumn{1}{l|}{Bicubic}  & $\surd$ & 4.28 & 7.14 & 11.58 & 2.28 & 3.98 & 6.37 & 2.42 & 4.54 & 7.38  \\
\multicolumn{1}{l|}{DG~\cite{Gu_2017_CVPR}} & \textbf{$\times$} & 3.68 & 5.78 & 10.08  & 1.97 & 4.16 & 5.27 & 2.06 & 4.19 & 6.90 \\
\multicolumn{1}{l|}{DJF~\cite{DJF}}   & \textbf{$\times$} & 3.54 & 6.20 & 10.21 & 2.14 & 3.77 & 6.12 & 2.54 & 4.71 & 7.66  \\
\multicolumn{1}{l|}{DJFR~\cite{DJFR}}  & \textbf{$\times$} & 2.38 & 4.94 & 9.18 & 1.32 & 3.19 & 5.57 & 1.15 & 3.57 & 6.77 \\
\multicolumn{1}{l|}{CUNet~\cite{CUNet}} & \textbf{$\times$} & 1.92 & 3.70 & 6.78 & 1.10 & 2.17 & 4.33 & 0.91 & 2.23 & 4.99  \\
\multicolumn{1}{l|}{PAC~\cite{su2019pixel}} & \textbf{$\times$} & 1.89 & 3.33 & 6.78& 1.32 & 2.62 & 4.58 & 1.20 & 2.33 & 5.19  \\
\multicolumn{1}{l|}{DMSG~\cite{DMSG2016}}& \textbf{$\times$}  & 3.02 & 5.38 & 9.17& 1.88 & 3.45 & 6.28 & 2.30 & 4.17 & 7.22  \\
\multicolumn{1}{l|}{FDKN~\cite{IJCV2021}} & \textbf{$\times$}  & 1.86 & 3.55 & 6.96& 1.09 & 2.17 & 4.51 & \underline{0.82} & 2.09 & 5.03  \\
\multicolumn{1}{l|}{DKN~\cite{IJCV2021}} & \textbf{$\times$}  & 1.62 & 3.26 & 6.51& 1.23 & 2.12 & 4.24 & 0.96 & 2.16 & 5.11  \\
\multicolumn{1}{l|}{FDSR~\cite{he2021towards}} & \textbf{$\times$} & 1.61 & 3.18 & 5.86 & 1.13 & 2.08 & 4.39 & 1.29 & 2.19 & 5.00  \\
\multicolumn{1}{l|}{DCTNet~\cite{DCTNet}} & \textbf{$\times$} & 1.59 & 3.16 & 5.84 & 1.10 & 2.05 & 4.19 & 0.88 & 1.85 & 4.39  \\
\multicolumn{1}{l|}{JIIF~\cite{JIIF}} & \textbf{$\times$} & \textbf{1.37} & 2.76 & 5.27 & 1.09 & 1.82 & 3.31 & 0.85 & 1.73 & 4.16  \\
\multicolumn{1}{l|}{AHMF~\cite{AHMF}} & \textbf{$\times$} & \underline{1.40} & 2.89 & 5.64 & 1.07 & \textbf{1.63} & \underline{3.14} & 0.88 & 1.66 & \textbf{3.71}  \\

\hline
\multicolumn{1}{l|}{GeoDSR-small (ours, channels=64)}& $\surd$ & 1.48 & \underline{2.73} & \underline{5.10} & \textbf{1.04} & 1.73 & 3.19 & \underline{0.82} & \underline{1.62} & 4.11 \\
\multicolumn{1}{l|}{GeoDSR (ours, channels=128)}& \textbf{$\surd$}  & 1.42 & \textbf{2.62} & \textbf{4.86}& \textbf{1.04} & \underline{1.68} & \textbf{3.10} & \textbf{0.81} & \textbf{1.59} & \underline{3.92}  \\
\bottomrule[ 2pt]
\end{tabular} 
\end{center}
\caption{Quantitative comparison with the state-of-the-art methods in terms of RMSE. The best performance is shown in \textbf{bold} and second best performance is the \underline{underscored} ones. Note that only GeoDSR and Bicubic support continuous representation, while other methods require training specialized models for each scale.}
\label{table:rmse}
\end{table*}
\setlength{\tabcolsep}{1.4pt}

\section{Experiments}

Our proposed method constructs continuous representation for depth maps under the guidance of RGB images. Given an arbitrary sampling matrix, the model queries the value of each coordinate and outputs the sampling result. When the coordinates in the sampling matrix are arranged regularly, the sampling degenerates into a normal depth map super-resolution task.

\subsection{Implementation details}
\subsubsection{Dataset}
We select three benchmark RGB-D datasets to evaluate the proposed framework:
 (1)  NYU V2 dataset~\cite{NYUv22012}. This dataset includes 1449 RGB-D image pairs captured by the Microsoft Kinect~\cite{kinect}. Following previous works, we use the first 1000 pairs for training and the rest 449 pairs for evaluation.
 (2) Lu dataset~\cite{Lu_dataset}. This dataset contains 6 RGB-D pairs acquired by the ASUS Xtion Pro camera~\cite{swoboda2014comprehensive}. It is used for testing.
 (3) Middlebury dataset~\cite{M_dataset1,M_dataset2}. This dataset contains 30 RGB-D pairs provided by \emph{Scharstein et al.} using the technique of structured light~\cite{1211354}. This dataset is also used for testing only.

 

\subsubsection{Two-Stage Training}
Since our task is more challenging and hard to train than previous works, direct training may lead to the non-convergence of the model. Thus, we adopt a two-stage training method.
In the first stage, we train a model with a fixed scale, and then we finetune this model in the second stage to learn a continuous representation. We randomly crop the original HR depth maps to $256\times256$ patches as ground truths and then use bicubic degradation with a scaling factor $s$ on them to generate LR inputs. 
In the first stage, $s$ is fixed at 8, and in the second stage, $s$ is randomly sampled from the uniform distribution $U(1,16)$. In order to avoid overfitting, all of these LR depth maps are normalized to $[0,1]$ before being fed into the networks. 

We use the L1 loss to measure the gap between the prediction results and the ground truths, and use the Adam optimizer~\cite{kingma2014adam} with $\beta_1 = 0.9$ and $\beta_2 = 0.99$ to update the parameters. In each stage, we set the initial learning rate as 0.0001, and then divide it by 0.2 every 60 epochs. In both stages, the model is trained with the first 1000 pairs of RGB-D images from the NYU v2 dataset for 200 epochs with a batch size of 1. 
We use a GeForce RTX 3090ti GPU to train the model, and the whole training process takes about 12 hours.

\begin{figure}[t]
\centering
\includegraphics[width=0.95\columnwidth]{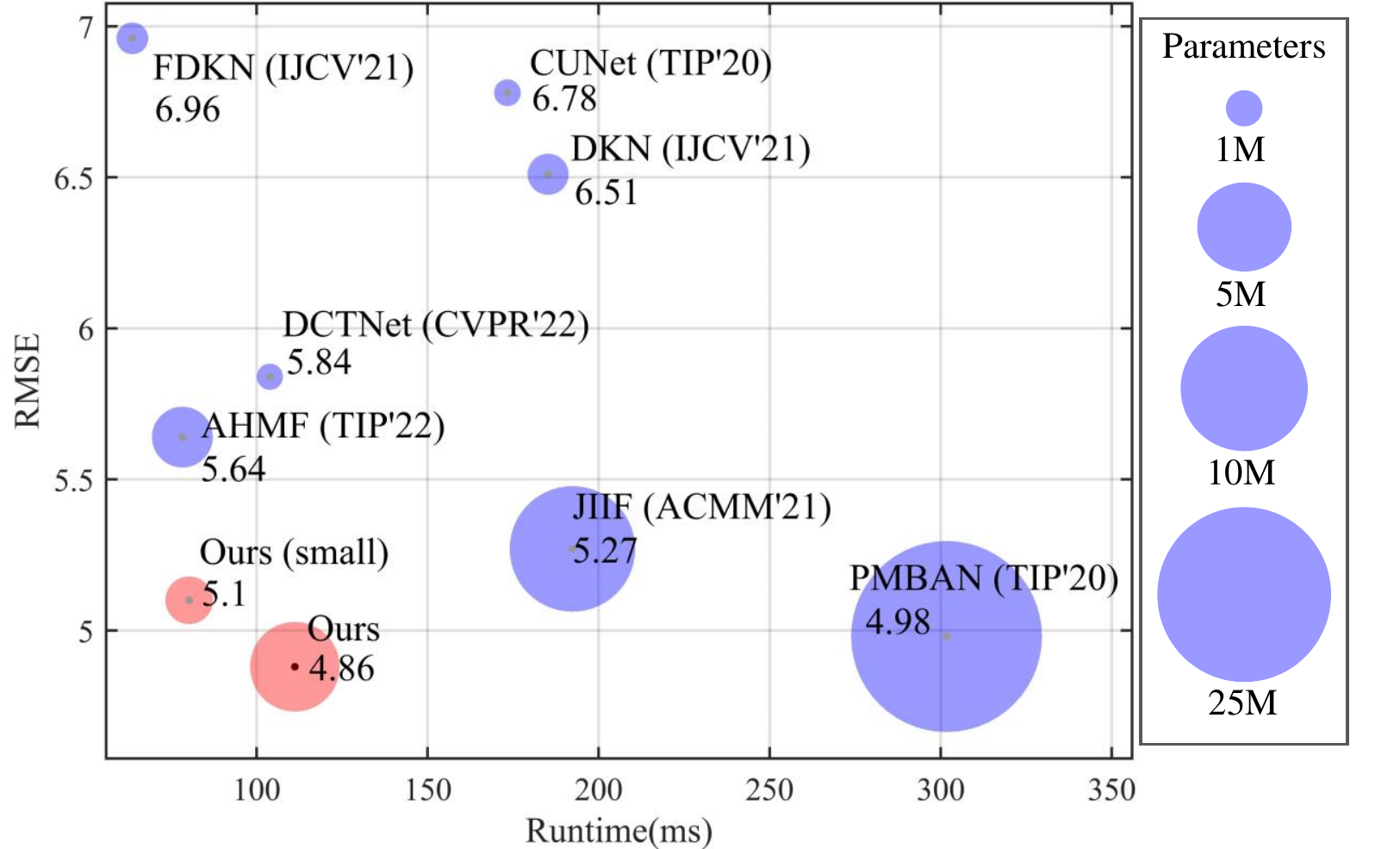}
\caption{Efficiency comparison of our method and other methods for $\times16$ GDSR on NYU v2 dataset. The corresponding RMSE (the lower, the better) results have been marked below the methods' name, and the circle's area represents the model's parameters.}
\label{fig:time}
\end{figure}

\begin{figure*}[ht]
\centering
\includegraphics[width=0.99\textwidth]{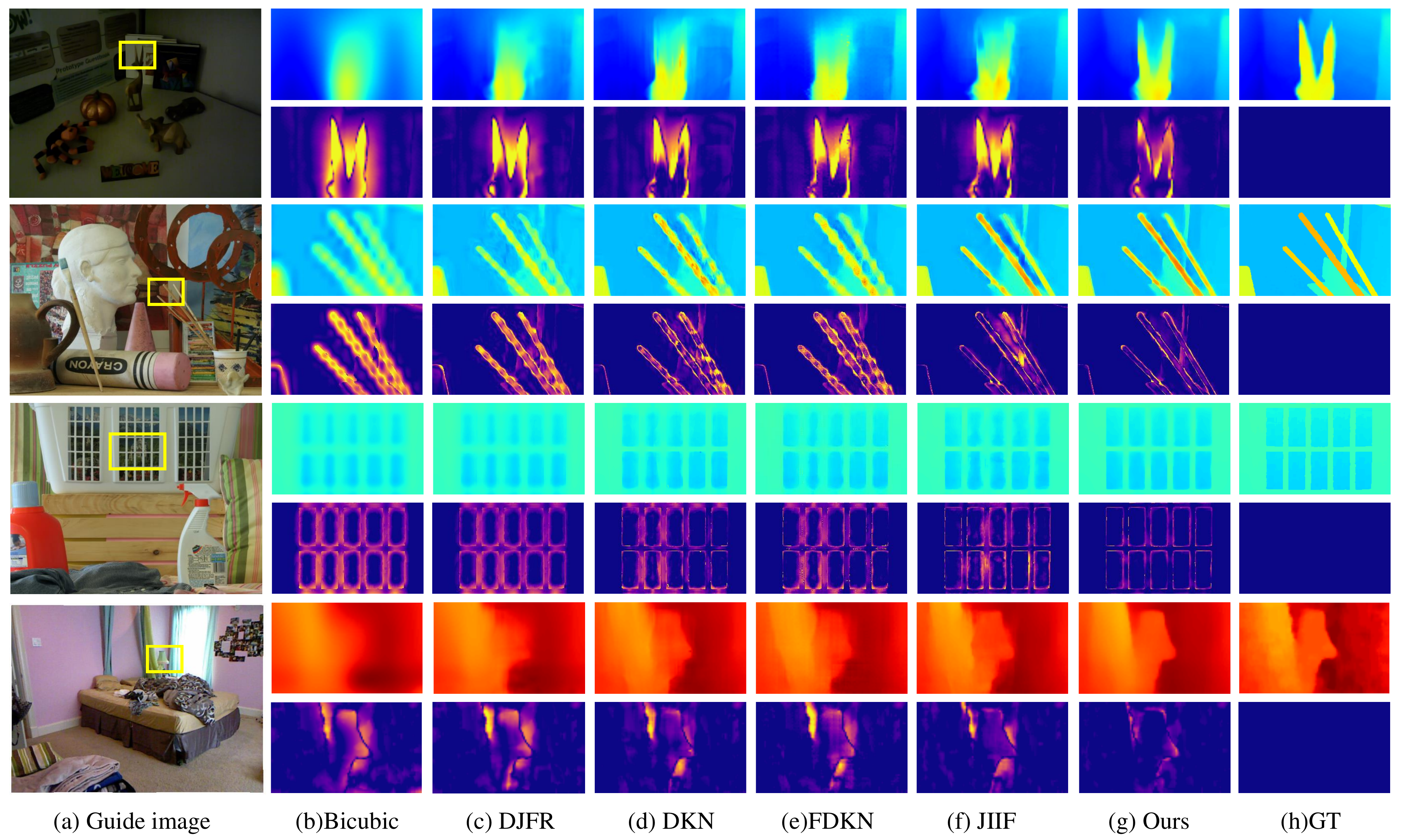}
\caption{Visual comparison of upsampled depth images ($\times16$).
For each sample, the second row shows the \textbf{error map} between the results and ground truth. The brighter area means the more significant error. More results can be found in our supplementary material. 
}
\label{fig:compare}
\end{figure*}

\subsection{Quantitative Comparison}



As shown in Table~\ref{table:rmse}, we test our model on the Middlebury dataset, Lu dataset, and NYU v2 dataset with the scale of $\times4$, $\times8$, and $\times16$. 
We also train a miniaturized GSA model with half of the channels to apply for scenarios with fewer computing resources. 
We use the average Root Mean Squared Error (RMSE) as the evaluation metric.
Following~\cite{JIIF}, the average RMSE is measured in centimeters for the NYU v2 dataset. As for the other two datasets, the depth values are scaled to [0, 255] to calculate RMSE since the source data is grayscale.

Since there are few works on the task of arbitrary-scale guided DSR to date, we have to draw comparisons with scale-fixed DSR methods. Note that all other methods in the table are trained and evaluated for a specific scale; thus, their task is much simpler than ours.
However, as Table~\ref{table:rmse} shows, our method outperforms them in most settings and is highly competitive in other cases, which fully demonstrates that GSA not only effectively constructs the continuous representation for depth maps but also efficiently utilizes the information of both guidance images and depth maps. Besides, it can be seen from Table~\ref{table:rmse} that the advantage of GSA becomes increasingly evident as the upsampling scale increases; this is because as the information provided by the depth map gradually decreases, the fusion of local feature perception and guidance information will become more critical. 
The efficiency of the proposed method is shown in Figure~\ref{fig:time}. Our model compares favorably against other approaches with relatively fewer parameters and quicker running speed. Considering that our model also has the advantage of arbitrary scale upsampling, the application value of our model is further demonstrated.

\subsection{Qualitative Comparison}

The visual comparison is shown in Figure~\ref{fig:compare}, where the input depth maps are upsampled by 16x to demonstrate more distinct differences. From top to bottom: Sample 1 is the results from the Lu dataset; Sample 2,3 are the results from the Middlebury dataset; and Sample 4 is from the NYU v2 dataset, respectively. Results of other upsampling scales can be found in our supplementary material. For each sample, the second row shows the error map between the results and ground truth, where a brighter area means the more significant error. Obviously, the results produced by our model are more precise, whose edges are also sharper.
These results illustrate that our method constructs accurate and detailed continuous representation for depth maps, producing visually far superior results to other methods.

\setlength{\tabcolsep}{2pt}
\begin{table}[t]
\begin{center}
\begin{tabular}{c|c|cccccc}
\hline
Method     & paras(M) & $\times3.75$  & $\times14.6$ & $\times20.25^*$ & $\times24^*$  & $\times30^*$  \\ \hline
Bicubic    & -        & 4.07    & 11.09 & 13.57   & 14.91 & 16.88 \\ \hline
JIIF$^\dag$      & 10.83    & 1.52    & 5.08  & 6.85    & 7.96  & 9.76  \\
Ours-s     & 1.56     & 1.41    & 4.79  & 6.51    & 7.64  & 9.68  \\
Ours       & 5.52     & 1.35    & 4.56  & 6.22    & 7.26  & 9.23  \\ \hline
\end{tabular}
\end{center}
\caption{RMSE of scale-continuous sampling on NYU v2 dataset.$^*$: The scale is out of the training distribution ($\times1$ to $\times16$).
$^\dag$: Please note that the original JIIF does not support continuous representation, and we retrain the model under our framework to extend it to arbitrary scale upsampling.}
\label{table:conti}
\end{table}

\begin{figure}[t]
\centering
\includegraphics[width=1\columnwidth]{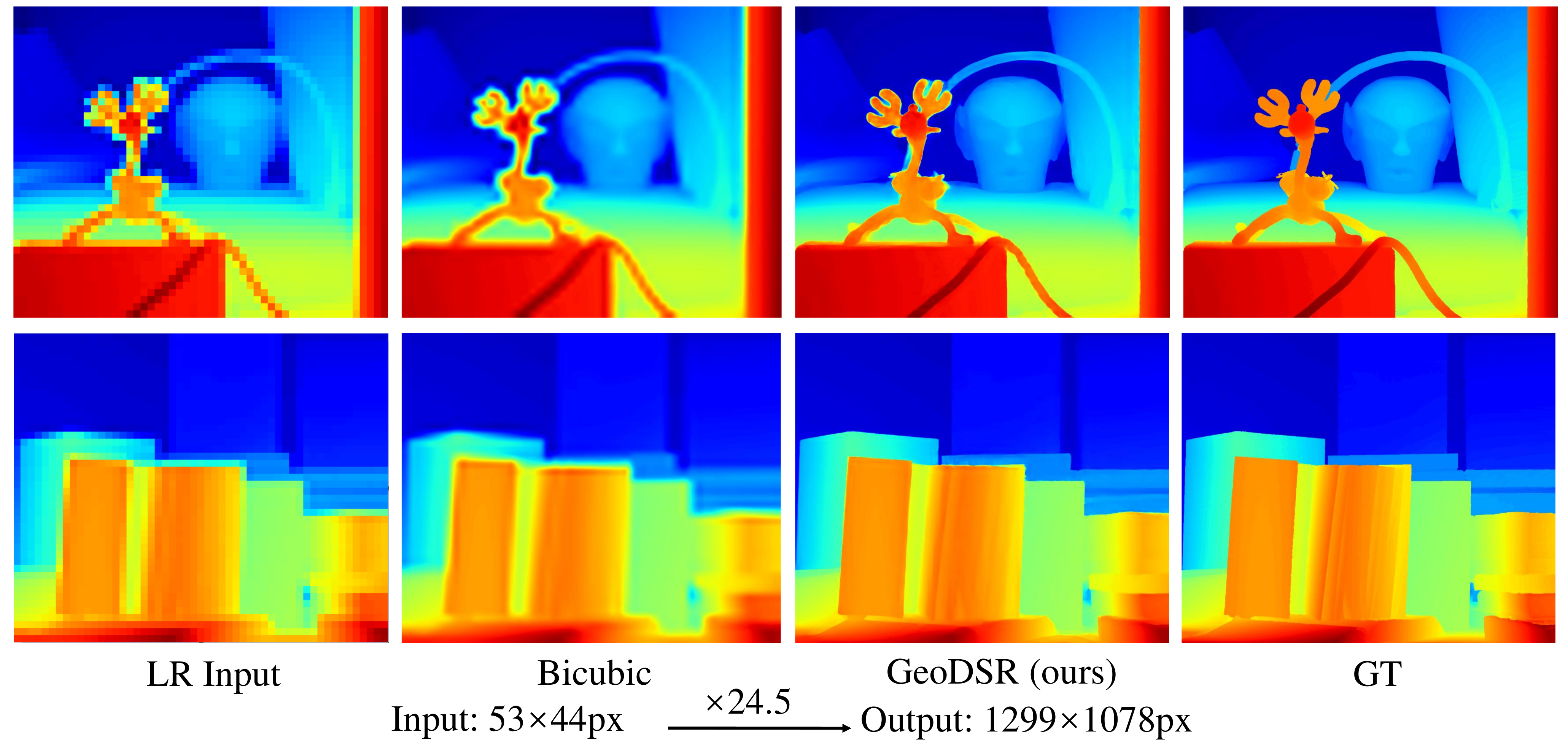}
\caption{Sample a $24.5\times$ higher resolution depth map from our continuous representation. The scale is non-integer and out of the training distribution.}
\label{fig:large}
\end{figure}

\subsection{Validation Experiments on Continuity}

\subsubsection{Scale-continuous Sampling}
As shown in Table~\ref{table:conti}, our model is able to upsample the depth map at any real-number scale even if the scale exceeds the training distribution ($\times1$ to $\times16$), which is the most critical function of our method. To draw a comparison, we retrain the model of JIIF~\cite{JIIF} (whose original version is implicit field based but does not support arbitrary scales) under our framework to extend it to arbitrary scale upsampling. The results show that our method has advantages in different real-number magnification within and outside the training distribution.
Figure~\ref{fig:large} shows an example of $\times24.5$ upsampling. The scale is non-integer and far beyond the training distribution, but our model still predicts highly clear results, which proves our method's continuity and strong generalization.

\begin{figure}[t]
\centering
\includegraphics[width=0.95\columnwidth]{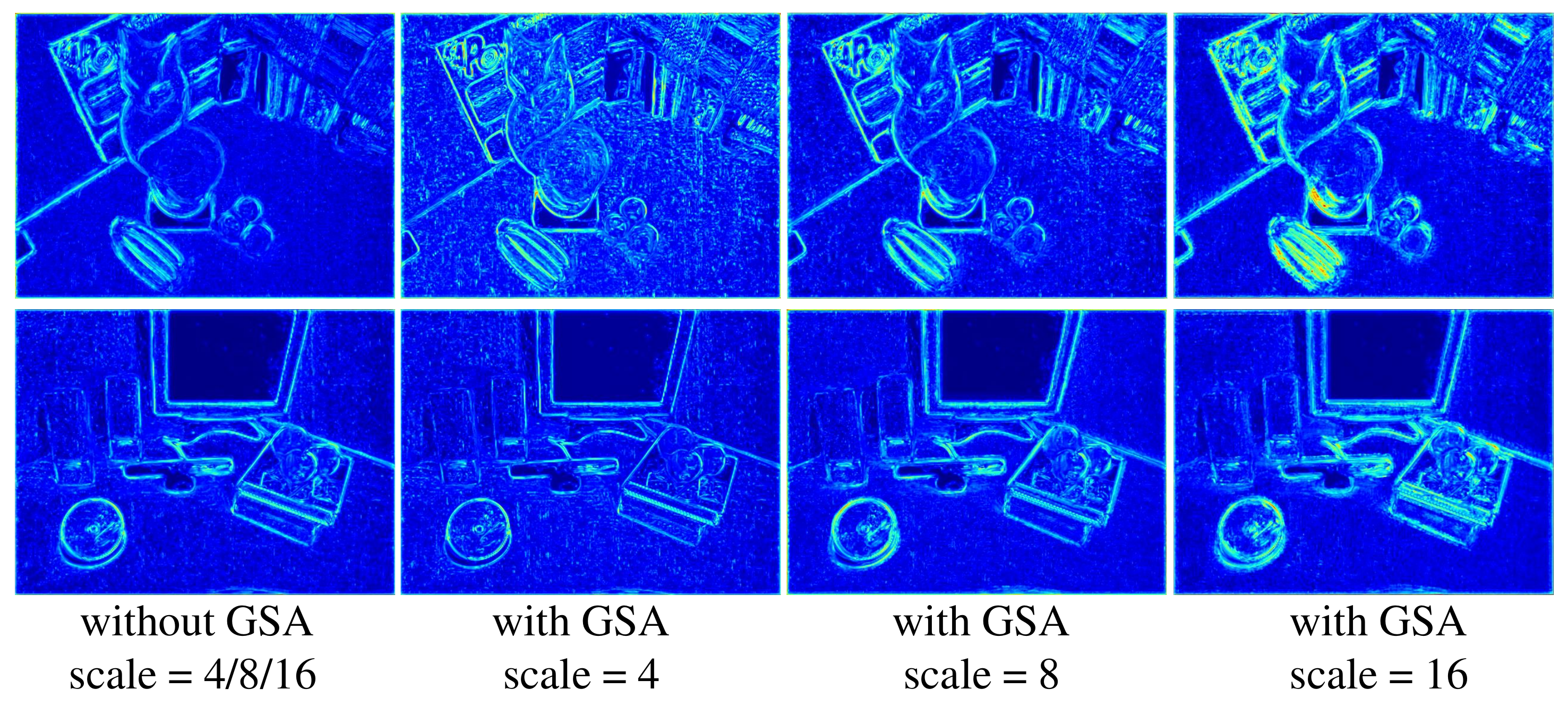}
\caption{Feature maps extracted from guidance images by our network at different scales with the same model. Brighter means stronger signal. 
The RGB image will provide more information around the edges to help construct the spatial geometry when the scale is larger.}
\label{fig:GASA_feature}
\end{figure}

\subsubsection{Features Extracted by GSA}
Figure~\ref{fig:GASA_feature} shows the RGB features extracted by our network under different conditions. With the same input, when GSA is disabled, the obtained RGB features are identical at all resolutions since the encoder module cannot receive the sampling's geometric information at all. After enabling GSA, it is evident that as the scale becomes larger, the signals are more and more concentrated near the crucial edges. This is because the larger the scaling factor, the less information the depth map provides, and there is greater uncertainty in the boundaries of the objects; therefore, the network needs the RGB image to provide more information around the edges to help construct the spatial geometry. This experiment fully proves that GSA can dynamically adjust the feature extraction according to the sampling conditions, thus effectively supporting the construction of continuous representations.

\begin{figure}[t]
\centering
\includegraphics[width=0.97\columnwidth]{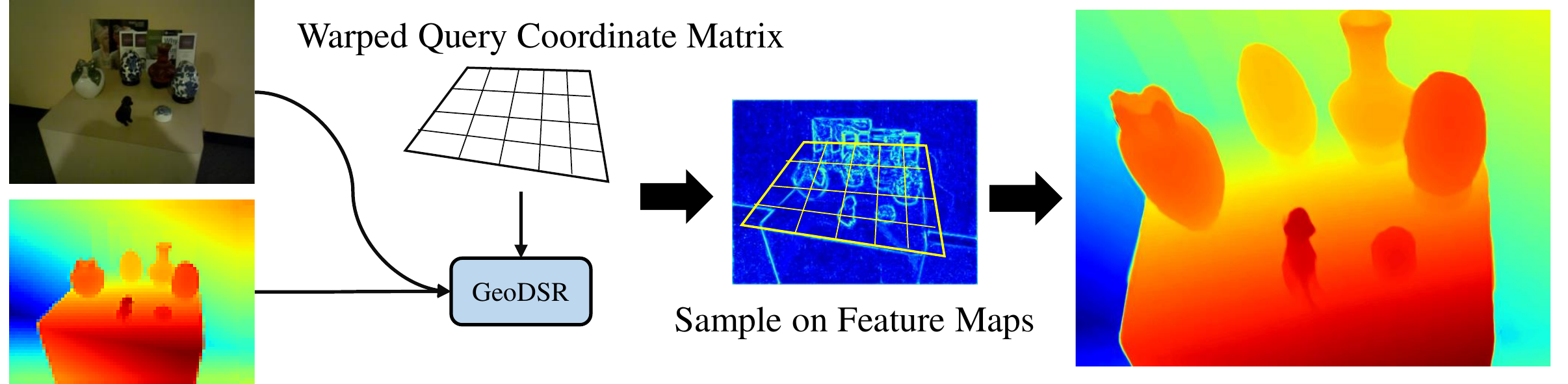}
\caption{Sample a warped image with GeoDSR. Give the model a sampling matrix of a particular shape (\emph{i.e.}, manual design or affine transformation of the typical rectangular sampling matrix), and a warped shape will be sampled.}
\label{fig:warp}
\end{figure}

\subsubsection{Spatial-continuous Sampling}
Our continuous representation can also be used to sample a warped image. As shown in Figure~\ref{fig:warp}, give GSA a specially designed coordinate matrix, and the model will return a warped result. 

\setlength{\tabcolsep}{2pt}
\begin{table}[t]
\begin{center}
\begin{tabular}{c|c|c|c|c|c}
\hline
 & Configurations         & paras(M) & $\times4$ & $\times8$ & $\times16$  \\ \hline
I     & w/o GSA               & 5.34     & 1.45 & 2.81 & 5.50 \\
II     & w/o feature modulation      & 5.71     & 1.45 & 2.67 & 4.91 \\
III     & w/o two-step upsampling  & 5.52   & 1.41 & 2.70 & 5.03 \\
IV     & w/o two-stage learning & 5.52     & 1.47 & 2.75 & 5.07 \\ \hline
V     & our full model         & 5.52     & 1.42 & 2.62 & 4.86 \\
\hline
\end{tabular}
\caption{Results of ablation studies on the NYU v2 test set. }
\label{table:ablation}
\end{center}
\end{table}

\subsection{Ablation study}  \label{ablation study}


\subsubsection{Effect of Geometry-Aware Spatial Aggregator}
The role of GSA is to extract the corresponding features according to different sampling conditions for subsequent upsampling.
In Model I, we replace the geometric perception convolution of GSA with ordinary convolution while maintaining other structures of the operator unchanged. 
As shown in Table~\ref{table:ablation}, the performance of the model is significantly reduced, especially at the scale of $\times16$ (5.50 vs. 4.86). This is because, in the absence of GSA, the model cannot dynamically extract the features according to different scales, thus only obtaining a compromised result. 
As shown in Figure~\ref{fig:GASA_feature}, the feature map of Model I (without GSA) is relatively more different from the $\times 16$ feature map of the entire model, so the results on $\times16$ degrade the most.

\subsubsection{Effect of Feature Modulation Module}
The function of the feature modulation module is to pre-mix the information of the guidance image and the depth map. We remove our modulation-based fusion function in model II and concatenate the features directly. Since the number of channels will increase after concatenation, the model's parameters also increase. However, as shown in Table~\ref{table:ablation}, Model II gets worse results with more parameters, which proves the effectiveness of our simple yet efficient fusion module.

\subsubsection{Effect of Other Designs}
We also analyze the effectiveness of the model's two-step upsampling mechanism and two-stage training strategy. In Model III, we disable the two-step upsampling mechanism and upsample the features to the final resolution at the first decoder. The results show that the performance of one-step sampling is obviously weaker than our method. In Model IV, we change the two-stage training strategy, cancel the first stage of learning fixed scale upsampling, and directly train an arbitrary scale model from scratch. The results show that our two-stage training method is more conducive to the model's convergence. 
Due to page limitation, studies of the other hyperparameters including the fixed scale in the first learning stage and the number of GSA groups are discussed in the supplementary material.

\section{Conclusion}
In this paper, we propose a contextualized continuous representation neural network GeoDSR for depth maps, with whose help we can achieve arbitrary forms of sampling (including arbitrary sampling scales and arbitrary sampling shapes). 
We design an operator named Geometric Spatial Aggregator (GSA) as the feature extractor and a simple yet efficient feature modulation upsampler as the decoder for learning this continuous representation.
The aggregators efficiently utilize information from both guidance images and depth maps by calculating the correlation between sampling scales and spatial geometries.
The feature modulation module realizes the pre-fusion of features and simplifies the task of multi-modal learning.
Comparison experiments demonstrate the effectiveness and practicality of our approach.


\section{Acknowledgments}
This work was supported by National Science Foundation of China (U20B207, 61976137). This work was also partially supported by Grant YG2021ZD18 from Shanghai Jiaotong University Medical Engineering Cross Research.

\bibliography{aaai23}

\bigskip

\end{document}